\documentclass[10pt,twocolumn,letterpaper]{article}

\usepackage{iccv}
\usepackage{times}
\usepackage{epsfig}
\usepackage{graphicx}
\usepackage{amsmath}
\usepackage{amssymb}

\usepackage{multirow}
\usepackage{bbm}
\usepackage[table]{xcolor}
\usepackage{arydshln}

\usepackage{subfigure}

\usepackage{listings}
\usepackage{algorithm}% http://ctan.org/pkg/algorithms
\usepackage{algorithmic}
\usepackage{setspace}
\DeclareMathAlphabet\mathbfcal{OMS}{cmsy}{b}{n}

\definecolor{mygreen}{HTML}{39b54a}  % green
\definecolor{myred}{HTML}{A10035}
\definecolor{myyellow}{HTML}{F8E924}
\definecolor{ggray}{RGB}{127,127,127}
\definecolor{ggreen}{HTML}{3EC70B}
\definecolor{mygray1}{gray}{.5}
\definecolor{mygray}{gray}{.9}
\definecolor{aliceblue}{rgb}{0.94, 0.97, 1.0}

\usepackage{pifont}
\newcommand{\cmark}{\textcolor{purple}{\ding{52}}}%

\definecolor{voc_cow}{HTML}{0C1E7F}
\definecolor{voc_horse}{HTML}{FB2576}
\definecolor{violet}{HTML}{BB1AEF}

\newcommand{\listnumber}[1]{\textbf{\color{violet}{#1}}}
\newcommand{\pub}[1]{\color{gray}{\tiny{#1}}}

\usepackage{tabulary}
\newcolumntype{I}{!{\vrule width 1pt}}

\newcolumntype{x}[1]{>{\centering\arraybackslash}p{#1pt}}
\newcolumntype{y}[1]{>{\raggedright\arraybackslash}p{#1pt}}
\newcolumntype{z}[1]{>{\raggedleft\arraybackslash}p{#1pt}}
\newlength\savewidth

\makeatletter
\newcommand{\thickhline}{%
	\noalign {\ifnum 0=`}\fi \hrule height 1pt
	\futurelet \reserved@a \@xhline
}
\makeatother

\newcommand{\increase}[1]{
	{\fontsize{7pt}{0.5em}\selectfont\color{purple}{$\uparrow$~{#1}}}
}

\newcommand{\deincrease}[1]{
	{\fontsize{7pt}{0.5em}\selectfont\color{ggreen}{$\downarrow$~{#1}}}
}

\newcolumntype{x}[1]{>{\centering\arraybackslash}p{#1pt}}
\newcolumntype{y}[1]{>{\raggedright\arraybackslash}p{#1pt}}
\newcolumntype{z}[1]{>{\raggedleft\arraybackslash}p{#1pt}}

\usepackage[misc]{ifsym}

\usepackage[pagebackref=true, breaklinks=true, letterpaper=true, colorlinks, citecolor=citecolor, linkcolor=linkcolor, bookmarks=false]{hyperref}
\definecolor{citecolor}{HTML}{0071BC}
\definecolor{linkcolor}{HTML}{ED1C24}

\iccvfinalcopy % *** Uncomment this line for the final submission

\def\iccvPaperID{****} % *** Enter the ICCV Paper ID here

\ificcvfinal\fi

\begin{document}

%%%%%%%%% TITLE
\title{Parallel Vertex Diffusion for Unified Visual Grounding}

\author{
Zesen Cheng$^{1}$\thanks{\Letter~Corresponding Author.} \quad Kehan Li$^{1}$ \quad Peng Jin$^{1}$ \quad Xiangyang Ji$^{3}$\\
Li Yuan$^{1,2}$ \quad Chang Liu$^{3}$~\textsuperscript{\Letter} \quad Jie Chen$^{1,2}$~\textsuperscript{\Letter}\and
$^{1}$ School of Electronic and Computer Engineering, Peking University \\
$^{2}$ Peng Cheng Laboratory \quad
$^{3}$ Tsinghua University \quad \\
% \small{\tt{wanli.ouyang@sydney.edu.au},  \tt{danxu@cse.ust.hk}}
}
% \authornote{\Letter Corresponding Author}
\renewcommand\footnotemark{}
\maketitle
\ificcvfinal\thispagestyle{empty}\fi

\begin{abstract}
Unified visual grounding pursues a simple and generic technical route to leverage multi-task data with less task-specific design.
The most advanced methods typically present boxes and masks as vertex sequences to model referring detection and segmentation as an autoregressive sequential vertex generation paradigm.
However, generating high-dimensional vertex sequences sequentially is error-prone because the upstream of the sequence remains static and cannot be refined based on downstream vertex information, even if there is a significant location gap.
Besides, with limited vertexes, the inferior fitting of objects with complex contours restricts the performance upper bound.
To deal with this dilemma, we propose a parallel vertex generation paradigm for superior high-dimension scalability with a diffusion model by simply modifying the noise dimension.
An intuitive materialization of our paradigm is Parallel Vertex Diffusion~(PVD) to directly set vertex coordinates as the generation target and use a diffusion model to train and infer.
We claim that it has two flaws: (1) unnormalized coordinate caused a high variance of loss value; (2) the original training objective of PVD only considers point consistency but ignores geometry consistency.
To solve the first flaw, Center Anchor Mechanism~(CAM) is designed to convert coordinates as normalized offset values to stabilize the training loss value. 
For the second flaw, Angle summation loss~(ASL) is designed to constrain the geometry difference of prediction and ground truth vertexes for geometry-level consistency.
Empirical results show that our PVD achieves state-of-the-art in both referring detection and segmentation, and our paradigm is more scalable and efficient than sequential vertex generation with high-dimension data.
\end{abstract}

\begin{figure}[t]
\centering
\includegraphics[width=1.0\linewidth]{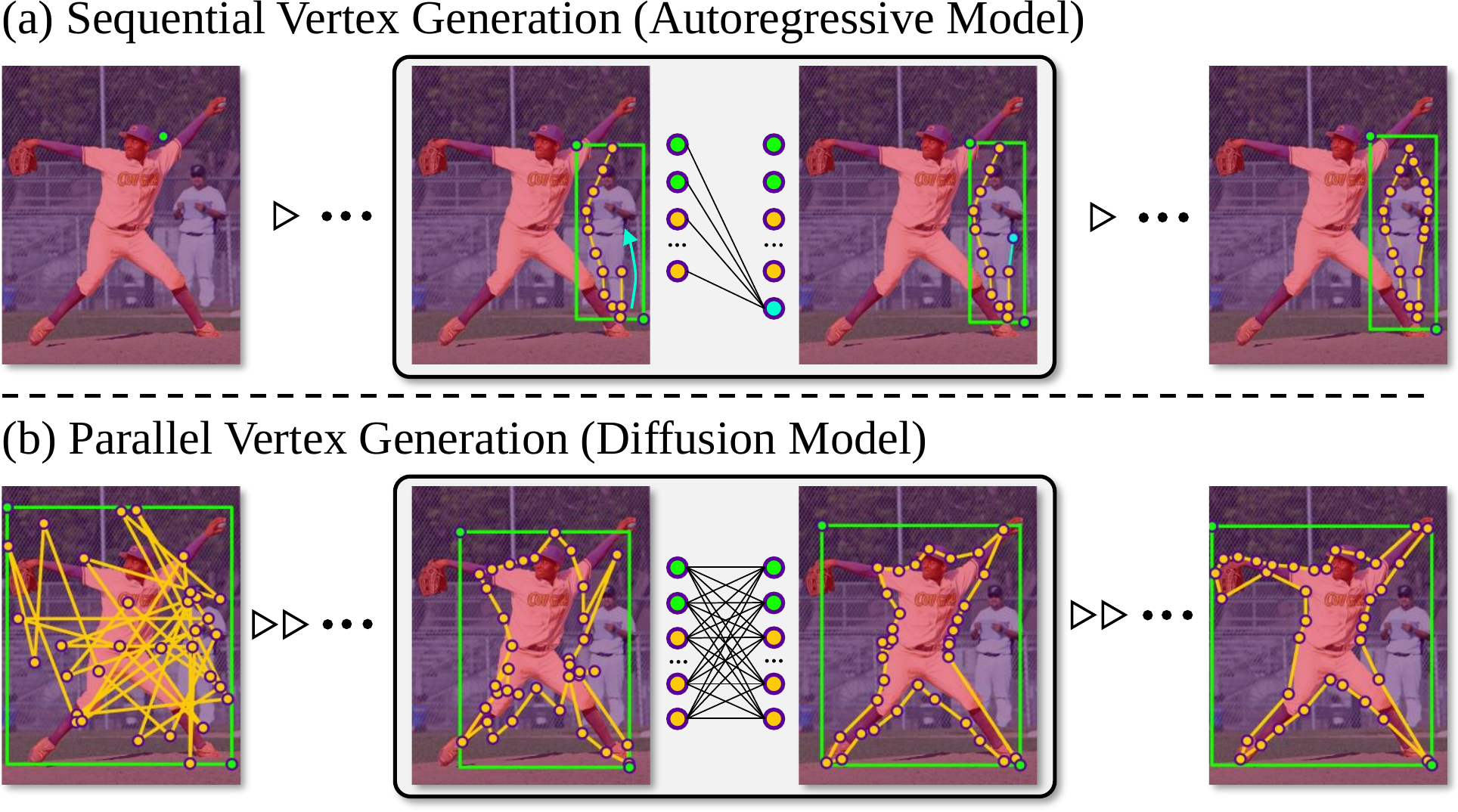} 
% SVG 的不好: 指数时间的序列长度放缩, 误差积累;
\caption{\textbf{Comparison} between \listnumber{(a)} the most advanced unified visual grounding paradigm, i.e., Sequential Vertexes Generation and \listnumber{(b)} our proposed paradigm, i.e., Parallel Vertexes Generation. Because upstream vertexes is stationary in the subsequent generation, the former is easily trapped in error accumulation if upstream vertexes doesn't hit the right object. the latter updates all of the vertexes in each round for conquering error accumulation. The related referring expression of this image is ``\textit{center baseman}".
}
\vspace{-10pt}
\label{fig:motivation}
\end{figure}
\begin{figure*}[t]
\centering
\includegraphics[width=1.0\linewidth]{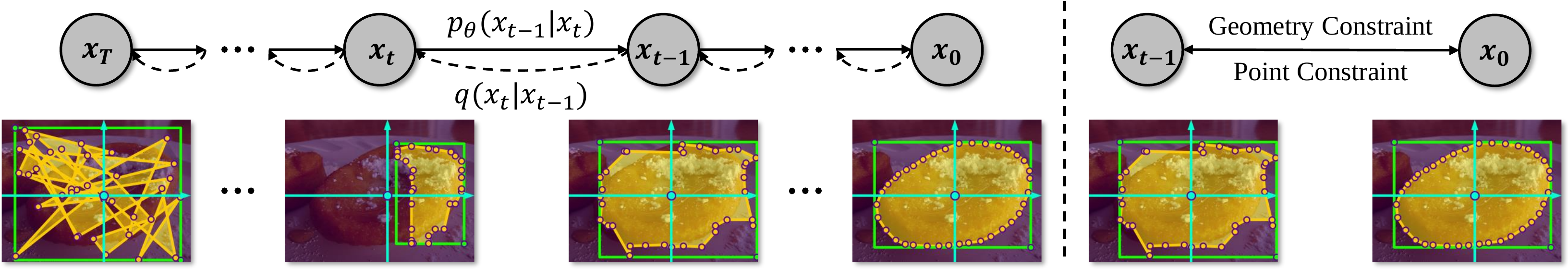} 
\caption{\textbf{Conceptual mechanism} of our Parallel Vertex Diffusion~(PVD) with Center Anchor Mechanism~(CAM) and Angle Summation Loss~(ASL). The vanilla PVD is hard to converge. 
CAM and ASL are extra designed for loss normalization and geometry consistency. 
}
\vspace{-10pt}
\label{fig:intuition}
\end{figure*}
\section{Introduction}

Visual grounding is an essential task in the field of vision-language that establishes a fine-grained correspondence between images and texts by grounding a given referring expression on an image~\cite{li2022grounded}. 
This task is divided into two sub-tasks based on the manner of grounding: Referring Expression Comprehension (REC) using bounding box-based methods~\cite{hu2016natural,mao2016generation,yu2016modeling}, and Referring Image Segmentation (RIS) using mask-based methods~\cite{hu2016segmentation}. 
REC and RIS are chronically regarded as separate tasks with different technology route, which requires complex task-specific design. However, REC and RIS share high similarities and have respective advantages so that it is natural and beneficial to unify two tasks. 
Recently, unified visual grounding becomes a trend of visual grounding because it avoids designing task-specific networks and can leverage the data of two tasks for mutual enhancement~\cite{luo2020multi}. 

The most advanced unified visual grounding paradigm represents boxes and masks as vertex sequences and models both REC and RIS as an autoregressive sequential vertex generation problem~\cite{zhu2022seqtr} which is solved by an autoregressive model~\cite{chen2021pix2seq}.
Although this paradigm unifies REC and RIS in a simple manner, it is hard to scale to high-dimension settings, which causes inferior fitting to objects with complex contours.
This flaw of sequential vertex generation paradigm mainly attributes to the sequential generation nature of its fundamental architecture, i.e., autoregressive model ~\cite{lin2020limitations,bond2021deep}.
Specifically, if upstream of vertex sequence is not precise enough, this paradigm easily gets stuck in a trap of error accumulation because upstream of sequence is stationary during subsequent generation.
In Fig.~\ref{fig:motivation}\textcolor{red}{(b)}, we can find that sequential vertex generation methods easily generate error vertexes when upstream vertexes don't hit the right object.
Moreover, sequential vertex generation methods face efficiency problem when scaling to high-dimension data because the iteration for generation sharply increases with the dimension of data~\cite{chen2022generalist}.

To tackle the issues of sequential vertex generation paradigm, we design parallel vertex generation paradigm to parallelly generate vertexes, which is more scalable for high-dimension data.
Specifically, the parallel nature allows dynamic modification of all vertexes to avoid error accumulation and requires only a few iterations to accomplish generation, which ensures it is easier to scale to high-dimension data.
Fig.~\ref{fig:motivation}\textcolor{red}{(b)} conceptually shows how our paradigm works.
Noisy vertexes are first sampled and then are gradually refined to precise vertexes of the object.

Recently, diffusion model~\cite{ho2020denoising} is demonstrated as an effective and scalable model when processing high-dimension generation tasks~(Text-to-Image generation~\cite{ramesh2022hierarchical}) and discriminative tasks~(Panoptic Segmentation~\cite{chen2022generalist}). 
The scalability of diffusion model mainly attributes to its parallel denoising mechanism. 
Diffusion model can easily scale to high-dimension settings by simply modifying the dimension of noise.
To leverage the scalability, we adopt a diffusion model to instantiate parallel vertex generation paradigm as Parallel Vertex Diffusion~(PVD) for more scalable unified visual grounding.
Moreover, we find that there are two aspects for improving the convergence of PVD: \listnumber{(1)} The coordinates of vertexes are not normalized so that the loss value has a high variance. \listnumber{(2)} The vanilla training objective of PVD only considers point-level consistency and ignores the geometry-level consistency between prediction and ground truth vertexes.
To implement this, we extra design Center Anchor Mechanism~(CAM) to normalize coordinate values for stabilizing the optimization signal and Angle Summation Loss~(ASL) to constraint geometry difference for achieving geometry consistency~(Fig.~\ref{fig:intuition}). 

Extensive experiments are conducted for both REC and RIS on three common datasets~(RefCOCO~\cite{yu2016modeling}, RefCOCO+~\cite{yu2016modeling} and RefCOCOg~\cite{mao2016generation}). On one hand, the empirical results show that PVD w/ CAM and ASL achieves SOTA on both REC and RIS tasks. On the other hand, the results also show that parallel vertex generation can better handle data with high-dimension and requires lower computation cost than sequential vertex generation paradigm.

\begin{figure*}[t]
\centering
\includegraphics[width=1.0\linewidth]{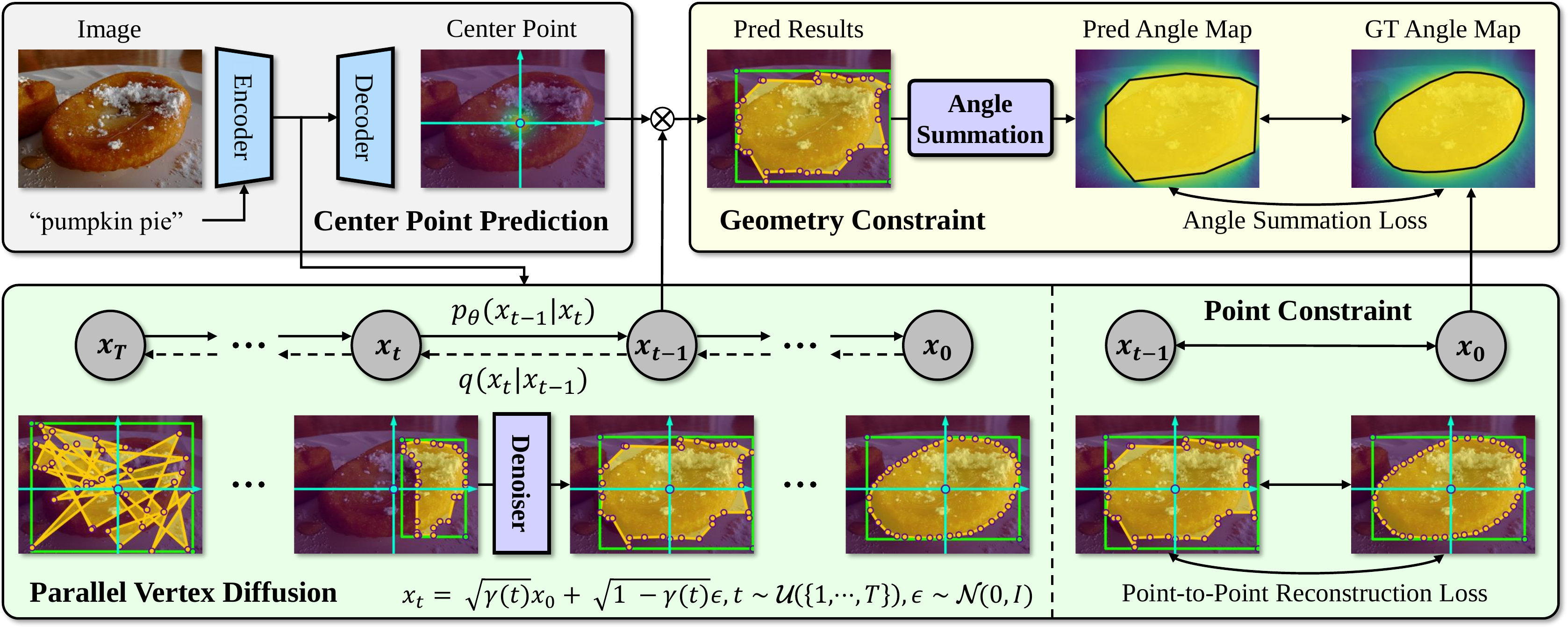} 
\caption{\textbf{Detailed workflow} has three components: \listnumber{(1)} \textit{Center Point Prediction} is used to extract cross-modal features and predict center point. \listnumber{(2)} \textit{Parallel Vertexes Diffusion} is used to generate normalized vertex vector by iteratively applying ``Denoiser" to noisy state $x_T$ and provides point level constraint during training phrase. After denormalized according to center anchor mechanism, the vertex vector is converted into vertex coordinates of bounding box and mask contour as prediction results. \listnumber{(3)} \textit{Geometry Constraint} utilizes angle summation algorithm to ensure the geometry consistency between prediction vertexes and ground truth vertexes for better optimization.
}
\vspace{-10pt}
\label{fig:framework}
\end{figure*}
\section{Related Work}

\subsection{Visual Grounding}

Visual grounding is essentially an object-centric fine-grained image-text retrieval~\cite{jin2022video,jin2022expectation,li2022grounded}, which has two forms: Referring Expression Comprehension~(REC) at box level~\cite{hu2016natural} and Referring Image Segmentation~(RIS) at mask level~\cite{hu2016segmentation}. 
Although they were parallel in the past, recent researches gradually focus on unified visual grounding because it requires little task-specific design and can leverage data from multiple tasks for mutual promotion.

\noindent\textbf{Referring Expression Comprehension.}
REC starts with the two-stage pipeline~\cite{hong2019learning,hu2017modeling,yu2018mattnet,liu2019learning,liu2019improving,luo2017comprehension,liu2017referring,yu2017joint,zhang2017discriminative,wang2019neighbourhood,zhang2018grounding,zhuang2018parallel,yang2019dynamic,chen2019referring}, where region proposals are first extracted by a pretrained detector~\cite{ren2015faster} and then ranked according to the reasoning similarity score with the referring expression. To tackle the two-stage pipeline's efficiency problem, researchers develop the one-stage pipeline to integrate location and cross-modal reasoning~\cite{li2022toward} into a single network for end-to-end optimization~\cite{liao2020real,yang2019fast,yang2020improving,sun2021iterative,huang2021look,deng2021transvg,zhou2021trar}.

\noindent\textbf{Referring Image Segmentation.}
RIS models the fine-grained referring object location as a pixel classification problem~\cite{hu2016segmentation}. The main focus of RIS research is to design better cross-modal alignment and fusion, e.g., concatenate~\cite{hu2016segmentation,li2018referring}, cross-modal attention mechanism~\cite{chen2019see,shi2018key,margffoy2018dynamic,ye2019cross,hu2020bi,ding2021vision,luo2020cascade}, visual reasoning~\cite{huang2020referring,hui2020linguistic,yang2021bottom}. Recent endeavors in RIS shift to simplify the pipeline of RIS system. EFN~\cite{feng2021encoder} couples the visual and linguistic encoder with asymmetric co-attention. LAVT~\cite{yang2021lavt} and ResTR~\cite{kim2022restr} replace those complex cross-modal attention and reasoning with a simple stack of transformer encoder~\cite{dosovitskiy2020image} layers.

\noindent\textbf{Unified Visual Grounding.}
Unified visual grounding has generally three types, i.e., two-stage paradigm~\cite{yu2018mattnet,liu2019learning,chen2019referring}, multi-branch paradigm~\cite{luo2020multi,li2021referring}, and sequential vertexes generation paradigm~\cite{zhu2022seqtr}. 
\listnumber{(1)} \textbf{Two-stage}: Methods in the two-stage paradigm are built upon a pretrained detector~(e.g., Mask R-CNN~\cite{he2017mask}) to first generate region proposals, and then referring expression is used to retrieve the most confident region as results~\cite{yu2018mattnet}. 
\listnumber{(2)} \textbf{Multi-task}: For breaking through the bottleneck of pretrained detector, multi-branch paradigm is proposed to assign task-specific branches to different tasks for joint end-to-end optimization~\cite{luo2020multi,li2021referring}. 
\listnumber{(3)} \textbf{Sequential Vertex Generation}: To simplify previous schemes and reduce optimization bias resulting from a multi-branch architecture~\cite{guo2020learning}, SeqTR models both REC and RIS as a vertex generation problem~\cite{zhu2022seqtr} and adopt an autoregressive model~\cite{chen2021pix2seq} to sequentially generate vertices of objects, which is currently the most advanced unified visual grounding scheme~\cite{zhu2022seqtr}.
Because of the sequential generation nature, SVG is easily trapped in error accumulation. This work proposes \textbf{Parallel Vertex Generation} to parallelly generate vertexes to avoid this issue.

\subsection{Generative Model for Perception}
 
Since Pix2Seq~\cite{chen2021pix2seq} first claims that sequence generation modeling is a simple and generic framework for object detection, generative model for perception is gradually gaining more attention. Following Pix2Seq, Pix2Seq v2~\cite{chen2022unified} is proposed as a general vision interface for multiple location tasks, e.g., object detection, instance segmentation, keypoint detection, and image caption. Although sequence generation modeling shows its potential, its fundamental generative architecture~(autoregressive model) hard scales to high-dimension data~\cite{chen2022generalist}. Subsequently, Pix2Seq-D shows that diffusion model~\cite{ho2020denoising,song2020denoising} is a better fundamental generative architecture for processing high-dimension task, i.e., panoptic segmentation~\cite{chen2022unified}. Then researchers expand diffusion model to other high-dimension tasks, e.g., semantic segmentation~\cite{amit2021segdiff,wolleb2022diffusion,graikos2022diffusion} and pose estimation~\cite{holmquist2022diffpose,choi2022diffupose}. 
% In this paper, we propose Parallel Vertex Diffusion for leveraging the capacity of diffusion model.

\section{Method}

\subsection{Overall Pipeline}

For claiming the relationship between different components of our workflow~(Fig.~\ref{fig:framework}), we describe the training and inference processes of the overall pipeline.

\noindent\textbf{Training.} An image-text pair $\{I, T\}$ is input into visual and linguistic encoder to extract cross-modal features $\mathbfcal{F}_c$~(Sec.\ref{sec:cfe}). The cross-modal features are then used to regress center point $c$~(Sec.\ref{sec:cam}). To create the ground truth of parallel vertex diffusion, center point is used to normalize the vertexes sampled from bounding box and mask contour for acquiring  ground truth vertex vector $\hat{V}$~(Sec.\ref{sec:pvd}). The final step is to calculate loss for optimization:
\begin{equation}
\label{eq:loss}
\mathcal{L} =  \mathcal{L}_c + \mathcal{L}_p + \mathcal{L}_g,
\end{equation}
where $\mathcal{L}_c$ is the center point loss~(Eq.~\ref{eq:ct_loss}) for optimizing center point prediction, $\mathcal{L}_p$ is the point-to-point reconstruction loss~(Eq.~\ref{eq:pt_loss}) for point level consistency, and $\mathcal{L}_g$ is the angle summation loss~(Eq.~\ref{eq:geo_loss}) for geometry consistency. The below sections will describe how to calculate these losses.

\noindent\textbf{Inference.} Same as the training process, we first predict the center point $c$~(Sec.\ref{sec:cam}). Then sampling a noisy state $x_T$ from standard gaussian distribution $\mathcal{N}(0, I)$. The noise is iteratively denoised to clean vertex vector $V_0$ by denoiser $f_{\theta}$~(Sec.\ref{sec:pvd}). Finally, denormalizing the coordinates of vertexes and convert the denormalized coordinates of vertexes to bounding box and binary mask by toolbox of COCO~\cite{lin2014microsoft}.

\subsection{Cross-modal Feature Extraction}
\label{sec:cfe}

The first step is to prepare cross-modal fusion features. An image-text pair $\{I\in \mathbb{R}^{h\times w\times3}, T\in\mathbb{N}^{n}\}$ is first sampled from dataset, where $h$, $w$, and $n$ denote the image height, the image width, and the number of words.
% Note that the text is preprocessed to word embeddings by Glove\cite{pennington2014glove}.
 Note that the words of text are already tokenized to related numbers by Bert tokenizer~\cite{wolf2020transformers}. 
To encoder image, the image is input in visual backbone~(e.g., ResNet~\cite{he2016deep}, Darnet53~\cite{redmon2018yolov3}, Swin~\cite{liu2021swin}) for acquiring  multi-scale visual features $\{\mathbf{F}_{v_1}\in\mathbb{R}^{\frac{h}{8}\times \frac{w}{8}\times d_1}, \mathbf{F}_{v_2}\in\mathbb{R}^{\frac{h}{16}\times \frac{w}{16}\times d_2},
\mathbf{F}_{v_3}\in\mathbb{R}^{\frac{h}{32}\times \frac{w}{32}\times d_3}\}$.
To encoder text, the word tokens of text are input in language backbone~(e.g., Bert~\cite{devlin2018bert}) for obtaining linguistic features $F_{l}\in\mathbb{R}^{n\times d_l}$. Specifically, the language backbone is the base bert model so that $d_l$ is $768$. Then we blend visual and linguistic features via element-wise multiplication~\cite{zhu2022seqtr} and adopt multi-scale deformable attention Transformer (MSDeformAttn)~\cite{zhu2020deformable} for multi-scale cross-modal fusion:
% a cross-modal FPN~\cite{lin2017feature,kirillov2019panoptic} is adopted to fuse visual features and linguistic features at multiple scales:
\begin{equation}
\mathbf{F}_{c_i} = \mathrm{MSDeformAttn}(\mathrm{MLP}(\sigma(\mathbf{F}_{v_i}) \odot \sigma(\mathbf{F}_{l}))),
\end{equation}
where $\odot$ denotes element-wise multiplication, $\mathrm{MLP}(\cdot)$ are three fully-connected layers and $\mathbf{F}_{c_i} \in \mathbb{R}^{h_i\times w_i\times d}$ denotes cross-modal features of $i$-th stage. $d$ is set to $256$.

\subsection{Center Anchor Mechanism}
\label{sec:cam}

The specific motivation of center anchor mechanism is to provide a coordinate anchor for converting coordinates to normalized offset values based on anchor, which is demonstrated to reduce the difficulty of coordinate regression~\cite{tian2019fcos}.

\noindent\textbf{Center Point Prediction.} 
the first step is to locate the rough center point $ c \in \mathbb{R}^2$ of the object referred by text. 
% Center point is the basic factor of center anchor mechanism.
Following previous advanced keypoint prediction network~(e.g., centernet~\cite{zhou2019objects}, cornernet~\cite{law2018cornernet}), the center point is represented as a gaussian heatmap $Y\in[0, 1]^{h\times w}$ by a gaussian kernel:
\begin{equation}
Y_{ij} = \mathrm{exp}(-\frac{(i - c_i)^2+(j - c_j)^2}{2\sigma^2}),
\end{equation}
where $\sigma^2$ is an image size-adaptive standard deviation~\cite{law2018cornernet}. Firstly, we use this point representation method to generate target gaussian heatmap $\overline{Y}$ of ground truth center point $\overline{c}$. Then the cross-modal features with largest spatial shape $\mathbf{F}_{v_1}$ are used to generate prediction heatmap $Y$, i.e., $Y = \mathrm{MLP}(\mathbf{F}_{v_1})$. Finally, a focal loss~\cite{lin2017focal} is adopted as the training objective for optimization:
\begin{equation}
\label{eq:ct_loss}
\mathcal{L}_c=\frac{1}{hw}\mathop{\sum}^{h,w}\limits_{i,j}\left\{\begin{aligned}
(1 - Y_{ij})^{\alpha}\mathrm{log}(Y_{ij}) & ,\overline{Y}_{ij} = 1, \\
(1 - Y_{ij})^{\beta}Y_{ij}^{\alpha}\mathrm{log}(1 - Y_{ij}) & ,\overline{Y}_{ij} < 1,
\end{aligned}
\right.
\end{equation}
where $\alpha$ and $\beta$ are hyper-parameters of the focal loss. Following previous works~\cite{law2018cornernet,zhou2019objects}, $\alpha$ and $\beta$ are set to $2$ and $4$, respectively. For resolving the prediction heatmap to concrete coordinate, we set the point with peak value of prediction heatmap as the prediction center point $c = \{i_c, j_c\}$.

\noindent\textbf{Coordinate Normalization.} To normalize a single point $p = \{i, j\}$, we adopt the scale and bias strategy:
\begin{align}
\hat{p} = (\hat{i}, \hat{j}) = (\frac{i - i_c}{h}, \frac{j - j_c}{w}),
\end{align}
where $\hat{p}$ denotes the point after normalizing.

\subsection{Parallel Vertex Diffusion}
\label{sec:pvd}

Parallel Vertex Diffusion~(PVD) is the key component for implementing parallel vertex generation paradigm. 
In this paper, we mainly refer to the usage of diffusion model in Pix2Seq-D~\cite{chen2022generalist}. The network which supports the reverse process of diffusion model is named ``Denoiser".

\begin{figure}[t]
\centering
\includegraphics[width=1.0\linewidth]{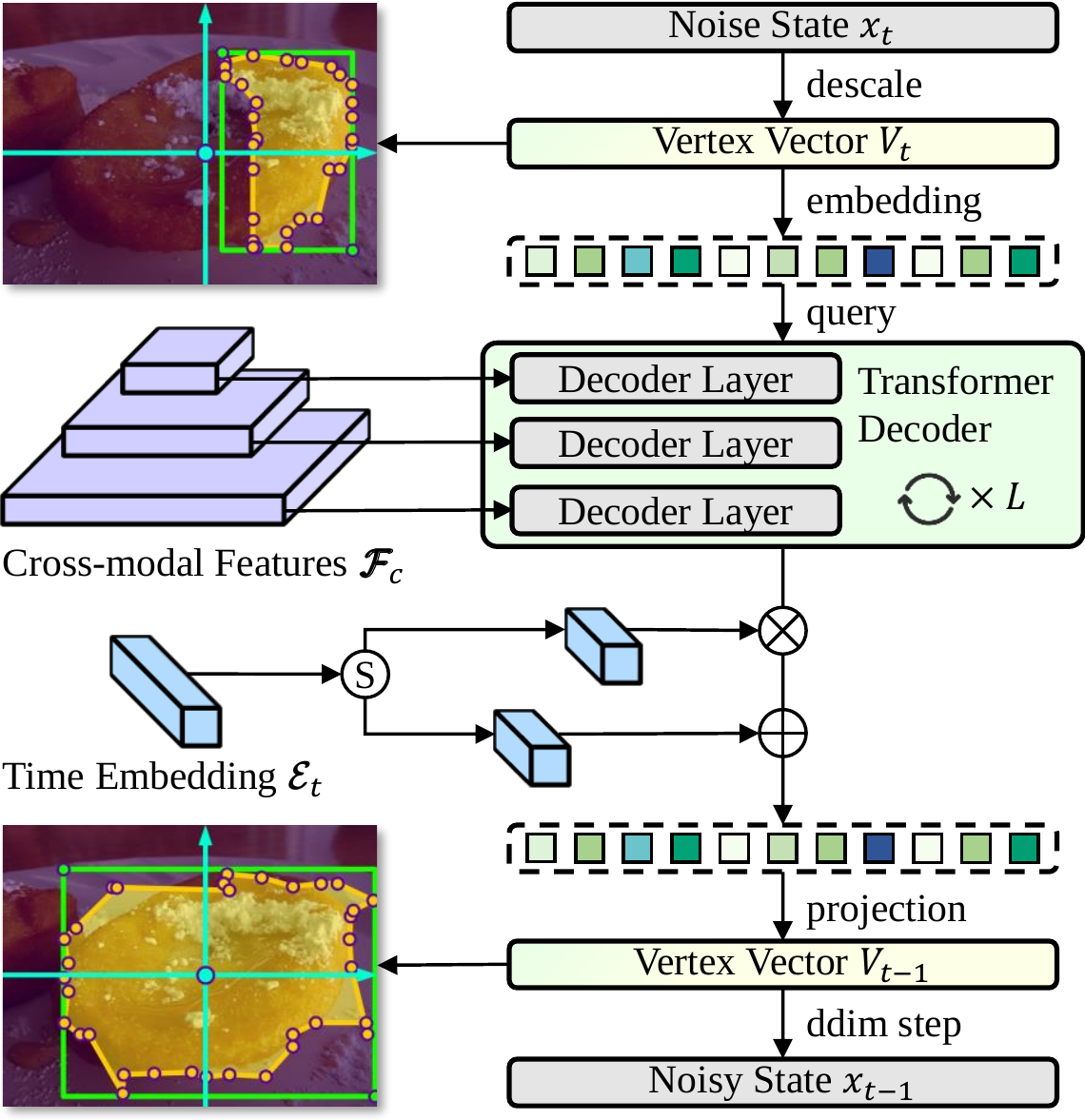} 
\caption{\textbf{Denoiser.} It is the parameterization network of diffusion model reverse process~$p_{\theta}(x_{t-1}|x_{t})$. Noisy state $x_t$ is descaled to vertex vector $V_t$ by Eq.~\ref{eq:descale}. The conversion tool from vertex vector to vertex embedding is the 2D coordinate embedding~\cite{meng2021conditional}. Refined vertex vector~$V_{t-1}$ and noisy state~$x_{t-1}$ are the output of Denoiser. The former is mainly used to get prediction results. The latter is mainly used to attend training objective~(Eq.~\ref{eq:pt_loss}).
}
\vspace{-10pt}
\label{fig:denoiser}
\end{figure}

\noindent\textbf{Denoiser.} Because normalized coordinates of vertexes are range in $[0, 1]$, it needs to be scaled to $[-b, b]$ for approximating gaussian distribution, where $b$ is set to $1$~\cite{chen2022analog}. Before inputting into denoiser, the noisy state $x_t$ needs to be clamped to $[-1, 1]$ and descaled from $[-1, 1]$ to $[0, 1]$:
\begin{equation}
\label{eq:descale}
V_t = \frac{(x_t / b + 1)}{2},
\end{equation}
where $x_t$ is the noisy state of $t$-th step, $V_t$ is the noisy vertex vector of $t$-th step. 
Denoiser is the denoise function for reverse transition of diffusion model. Firstly, the Denoiser converts the noisy state $x_t$ to vertex vector $V_t$ by Eq.~\ref{eq:descale}. Then the vertex vector is embedded to vertex embeddings $\mathcal{Q}_t$ by 2D coordinate embedding~\cite{meng2021conditional}. The embeddings are input into transformer decoder~\cite{carion2020end} as queries and interacted with cross-modal features $\mathbfcal{F}_c$. For keeping high-resolution features, three scales of cross-modal features are circularly utilized by different layer of the transformer decoder~\cite{cheng2021masked}. The circle is repeated $L$ times. After transformer decoder, the embedding is normalized by time embedding $\mathcal{E}_t$ and is projected to refined vertex vector $V_{t-1}$. Finally, The refined vertex vector is processed by DDIM step~\cite{song2020denoising} for acquiring  next noisy state $x_{t-1}$. To better understand the Denoiser, the detailed workflow is illustrated in Fig.~\ref{fig:denoiser}.

\noindent\textbf{Training phrase.} Suppose that we already sample a single image $I$ and its corresponding box $B\in\mathbb{R}^{2\times 2}$ and mask $ M\in\{0, 1\}^{h\times w}$ from dataset. \textcolor{violet}{\ding{182}} \textit{Preparation}: The first step is to extract and normalize the vertexes of the box and mask. The vertexes of box are left top corner and right bottom corner $\{p_b^{lt} = (i_b^{lt}, j_b^{lt}), p_b^{rb}=(i_b^{lt}, j_b^{rb})\}$. To represent mask as point sets, we retrieve mask contour via a classical contour detection algorithm~\cite{suzuki1985topological} implemented by opencv~\cite{bradski2000opencv}. Then the mask contour is used to sample the vertexes of mask $\{p_m^{1} = (i_m^{1}, j_m^{1}), \cdots, p_m^{N}=(i_m^{N}, j_m^{N})\}$, where $N$ is the sampling number of mask vertexes. The vertex set is finally normalized by center anchor mechanism~(Sec.~\ref{sec:cam}) and flattened to vertex vector ($\hat{V}\in\mathbb{R}^{4+2N}$): 
\begin{equation}
\hat{V}=\{\hat{i}_b^{lt}, \hat{j}_b^{lt}, \hat{i}_b^{rb}, \hat{j}_b^{rb}, \hat{i}_m^{1}, \hat{j}_m^{1}, \cdots, \hat{i}_m^{N}, \hat{j}_m^{N}\}.
\end{equation}
\textcolor{violet}{\ding{183}} \textit{Forward transition}: The normalized vertex vector $\hat{V}$ is set as the initial state $x_0$ of diffusion model and is forward transitioned to noisy state $x_t$, i.e., $q(x_t|x_0)$: 
\begin{equation}
\label{eq:diffusion}
x_t = \sqrt{\gamma(t)}x_0 + \sqrt{1 - \gamma(t)}\epsilon,
\end{equation}
where $\epsilon$ and $t$ are drawn from the standard normal distribution $\mathcal{N}(0, I)$ and uniform distribution $\mathcal{U}(\{1,\cdots,T\})$, $\gamma(t)$ denotes a monotonically decreasing function from 1 to 0. 
\textcolor{violet}{\ding{184}} \textit{Point constraint}: For training the denoiser, the noisy state $x_t$ is reversely transitioned to the noisy state $x_{t-1}$, i.e., $p_{\theta}(x_{t-1}|x_t)$. Following Bit Diffusion~\cite{chen2022analog}, the noisy state $x_{t-1}$ is required to approach initial state $x_0$:
\begin{equation}
\label{eq:pt_loss}
\mathcal{L}_p =  \mathbb{E}_{t\sim\mathcal{U}(\{1,\cdots,T\}),\epsilon\sim\mathcal{N}(0, I)}\Vert f_{\theta}(x_t, \mathbfcal{F}_c,t) - x_0 \Vert^2,
\end{equation}
where $f_{\theta}(\cdot)$ is the parameterized Denoiser, $x_t$ is derived by Eq.~\ref{eq:diffusion} from $x_0$, $\mathbfcal{F}_c=\{\mathbf{F}_{c_1}, \mathbf{F}_{c_2}, \mathbf{F}_{c_3}\}$ denotes all of the cross-modal features mentioned in Sec.~\ref{sec:cfe}. Eq.~\ref{eq:pt_loss} is point-to-point reconstruction loss.

\noindent\textbf{Inference phrase.} To generate vertex vector of box and mask, it requires a series of state transitions $x_{T}\rightarrow \cdots \rightarrow x_t \rightarrow \cdots \rightarrow x_{0}$. Because $T$ is set to a large value, $x_T\in\mathbb{R}^{4+2N}$ follows standard normal distribution $\mathcal{N}(0, I)$. Specifically, we sample a noise $x_T$ from standard normal distribution $\mathcal{N}(0, I)$ and iteratively apply Denoiser to denoise $x_T$ to implement the state transition chain for generation. During the generation, the vertex coordinates contained in noise vector $x_T$ are parallelly denoised to ground truth vertex coordinates, fully reflecting the characteristics of parallel generation.

\begin{figure}[t]
\centering
\subfigtopskip=2pt
\subfigbottomskip=5pt
\subfigcapskip=0pt
\begin{minipage}[b]{0.495\linewidth}
\subfigure[inside the polygon ($=360^\circ$)]{
    \includegraphics[width=1\linewidth]{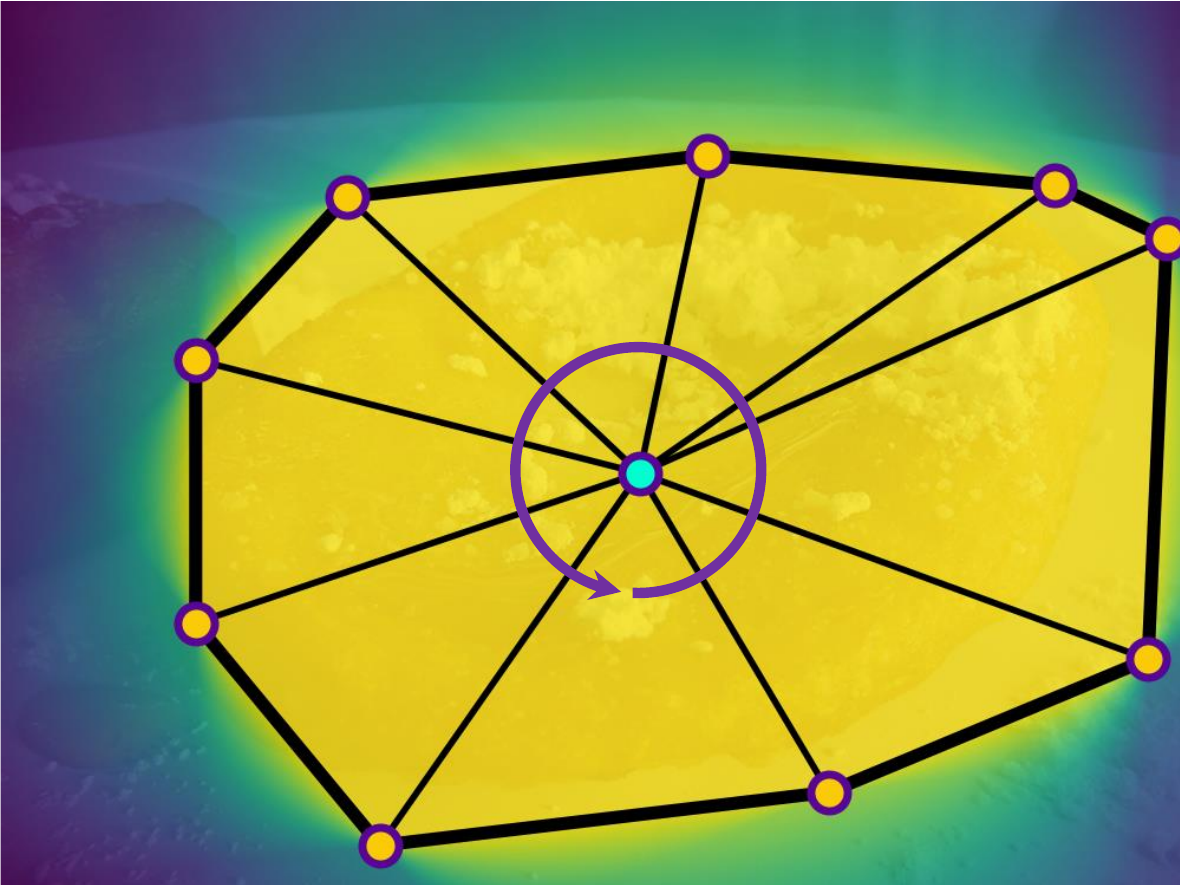}
    \label{fig:angle_summation_inner}
}
\end{minipage}
\hfill
\begin{minipage}[b]{0.495\linewidth}
\subfigure[outside the polygon ($<360^\circ$)]{
    \includegraphics[width=1\linewidth]{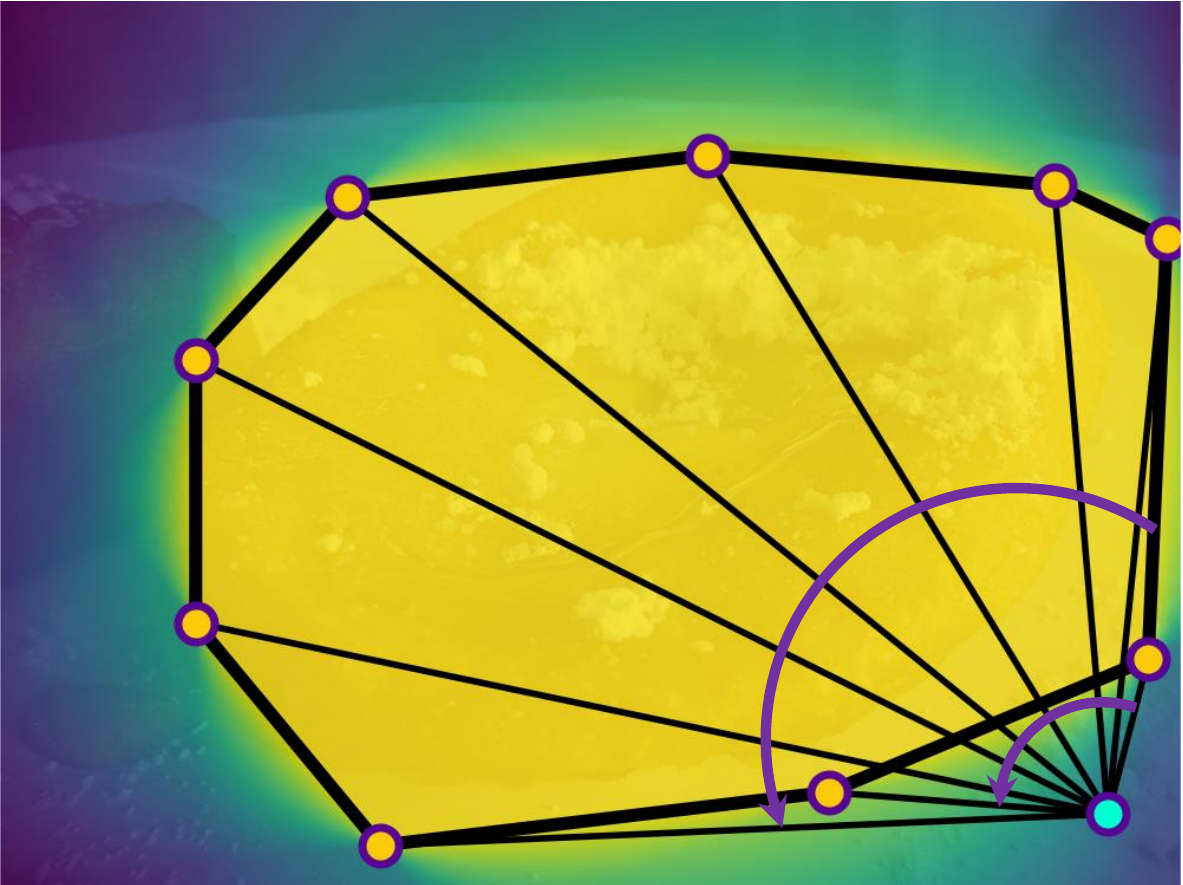}
    \label{fig:angle_summation_outer}
}
\end{minipage}
\caption{\textbf{Angle Summation.} \listnumber{(a)} If a point is inside the polygon, the sum of the angles between this point and each pair of points making up the polygon is $360^\circ$. \listnumber{(b)} If a point is outside the polygon, the sum of the angles is $<360^\circ$. According to the relative position between points and polygon, we can calculate an angle distribution on $h\times w$ spatial map, i.e., angle summation map.
% To clarify how to calculate angle summation map, we illustrate two cases of point location: 
}
\vspace{-10pt}
\end{figure}

\begin{table*}[t]
\centering
\renewcommand{\arraystretch}{1.3}
\setlength{\tabcolsep}{3mm}
\footnotesize
\begin{tabular}{z{80}|c|lll|lll|lll}
\noalign{\hrule height 1.5pt}
\multicolumn{1}{c|}{\multirow{2}{*}{Method}} & \multicolumn{1}{c|}{\multirow{2}{*}{Backbone}} & \multicolumn{3}{c|}{RefCOCO} & \multicolumn{3}{c|}{RefCOCO+} & \multicolumn{3}{c}{RefCOCOg} \\
\cline{3-11}
& & val & test A & test B & val & test A & test B & val(U) & test(U) & val(G) \\
\hline
\rowcolor{ggray!20}
\multicolumn{11}{c}{\it{Single-task}}\\
\hline
LBYL~\pub{CVPR2021}{\cite{huang2021look}} & Darknet53 & 79.67 & 82.91 & 74.15 & 68.64 & 73.38 & 59.49 & - & - & 62.70 \\
TransVG~\pub{CVPR2021}{\cite{deng2021transvg}} & ResNet101 & 81.02 & 82.72 & 78.35 & 64.82 & 70.70 & 56.94 & 68.67 & 67.73 & 67.02 \\
TRAR~\pub{ICCV2021}{\cite{zhou2021trar}} & Darknet53 & - & 81.40 & \underline{78.60} & - & 69.10 & 56.10 & 68.90 & 68.30 & - \\
\hline
\rowcolor{ggray!20}
\multicolumn{11}{c}{\it{Multi-task}}\\
\hline
MAttNet~\pub{CVPR2018}{\cite{yu2018mattnet}} & MRCNN-Res101 & 76.65 & 81.14 & 69.99 & 65.33 & 71.62 & 56.02 & 66.58 & 67.27 & - \\
NMTree~\pub{ICCV2019}{\cite{liu2019learning}} & MRCNN-Res101 & 76.41 & 81.21 & 70.09 & 66.46 & 72.02 & 57.52 & 65.87 & 66.44 & - \\
MCN~\pub{CVPR2020}{\cite{luo2020multi}} & Darknet53 & 80.08 & 82.29 & 74.98 & 67.16 & 72.86 & 57.31 & 66.46 & 66.01 & -\\
SeqTR~\pub{ECCV2022}{\cite{zhu2022seqtr}} & Darknet53 & 81.23 & 85.00 & 76.08 & 68.82 & 75.37 & 58.78 & \underline{71.35} & \underline{71.58} & -\\
\hdashline
\rowcolor{aliceblue!60} PVD & Darknet53 & \underline{82.51} & \underline{86.19} & 76.81 & \underline{69.48} & \underline{76.83} & \underline{59.68} & 68.40 & 69.57 & \underline{67.29} \\
\rowcolor{aliceblue!60} PVD & Swin-base & \textbf{84.52} & \textbf{87.64} & \textbf{79.63} & \textbf{73.89} & \textbf{78.41} & \textbf{64.25} & \textbf{73.81} & \textbf{74.13} & \textbf{71.51} \\
\noalign{\hrule height 1.5pt}
\end{tabular}
\vspace{3pt}
\caption{\textbf{Main results} on classical REC datasets. \textbf{Bold} denotes the best performance. \underline{Underline} denotes the second best performance.
}
\vspace{-5pt}
\label{tab:main_res_det}
\end{table*}
\begin{table*}[t]
\centering
\renewcommand{\arraystretch}{1.3}
\setlength{\tabcolsep}{3mm}
\footnotesize
\begin{tabular}{z{80}|c|lll|lll|lll}
\noalign{\hrule height 1.5pt}
\multicolumn{1}{c|}{\multirow{2}{*}{Method}} & \multicolumn{1}{c|}{\multirow{2}{*}{Backbone}} & \multicolumn{3}{c|}{RefCOCO} & \multicolumn{3}{c|}{RefCOCO+} & \multicolumn{3}{c}{RefCOCOg} \\
\cline{3-11}
& & val & test A & test B & val & test A & test B & val(U) & test(U) & val(G) \\
\hline
\rowcolor{ggray!20}
\multicolumn{11}{c}{\it{Single-task}}\\
\hline
CRIS~\pub{CVPR2022}{\cite{wang2022cris}} & CLIP-Resnet50 & 69.52 & 72.72 & 64.70 & 61.39 & 67.10 & 52.48 & 59.87 & 60.36 \\
LAVT~\pub{CVPR2022}{\cite{yang2022lavt}} & Swin-base & 72.73 & 75.82 & 68.79 & 62.14 & \underline{68.38} & 55.10 & 61.24 & 62.09 & \underline{60.50} \\
CoupleAlign~\pub{Neurips2022}{\cite{zhang2022coupalign}} & Swin-base & \underline{74.70} & \textbf{77.76} & \textbf{70.58} & \underline{62.92} & 68.34 & \underline{56.69} & \underline{62.84} & \underline{62.22} & - \\
\hline
\rowcolor{ggray!20}
\multicolumn{11}{c}{\it{Multi-task}}\\
\hline
MAttNet~\pub{CVPR2018}{\cite{yu2018mattnet}} & MRCNN-Res101 & 56.51 & 62.37 & 51.70 & 46.67 & 52.39 & 40.08 & 47.64 & 48.61 &      \\
NMTree~\pub{ICCV2019}{\cite{liu2019learning}} & MRCNN-Res101 &  56.59 & 63.02 & 52.06 & 47.40 & 53.01 & 41.56 & 46.59 & 47.88 & -     \\
MCN~\pub{CVPR2020}{\cite{luo2020multi}} & Darknet53 & 62.44 & 64.20 & 59.71 & 50.62 & 54.99 & 44.69 & 49.22 & 49.40 & - \\
SeqTR~\pub{ECCV2022}{\cite{zhu2022seqtr}} & Darknet53 & 67.26 & 69.79 & 64.12 & 54.14 & 58.93 & 48.19 & 55.67 & 55.64 & -\\
\hdashline
\rowcolor{aliceblue!60} PVD & Darknet53 & 68.87 & 70.53 & 65.83 & 54.98 & 60.12 & 50.23 & 57.81 & 57.17 & 54.15\\
\rowcolor{aliceblue!60} PVD & Swin-base & \textbf{74.82} & \underline{77.11} & \underline{69.52} & \textbf{63.38} & \textbf{68.60} & \textbf{56.92} & \textbf{63.13} & \textbf{63.62} & \textbf{61.33} \\
\noalign{\hrule height 1.5pt}
\end{tabular}
\vspace{3pt}
\caption{\textbf{Main results} on classical RIS datasets. \textbf{Bold} denotes the best performance. \underline{Underline} denotes the second best performance.
}
\vspace{-6pt}
\label{tab:main_res_seg}
\end{table*}

\subsection{Geometry Constraint}
\label{sec:asl}

The main optimization constraint of vertex generation is point-to-point reconstruction loss (Eq.~\ref{eq:pt_loss}). This loss can modify the prediction vertex set to one by one approach the ground truth vertex set and does not consider the geometry difference between two sets. Inspired by point-in-polygon problem of computational geometry, we introduce a simple geometry algorithm~(Angle Summation algorithm~\cite{sutherland1974characterization}) to make an optimization objective. 

\noindent\textbf{Angle Summation.} Computing the sum of the angles between the test point and each pair of points making up the polygon. As shown in \ref{fig:angle_summation_inner}, if this sum is equal to $360^\circ$ then the point is inside the polygon. As shown in \ref{fig:angle_summation_outer}, if this sum is less than $360^\circ$ then the point is outside the polygon. Therefore, we can calculate an angle summation map $\mathcal{A}\in(0, 360]^{h\times w}$ according to the relative position between each pixel and vertexes of polygon $V$. The details of how to convert coordinates of polygon vertexes to angle summation polygon please refer to the appendix.

\noindent\textbf{Training Objective.} According to the "Angle Summation", prediction vertexes $V$ and ground truth vertexes $\hat{V}$ are converted to prediction angle summation map $\mathcal{A}$ and ground truth angle summation map $\hat{\mathcal{A}}$. The angle summation map reflects the global geometry of vertex vector so that we hope the prediction angle summation map to approach the ground truth angle summation for reducing geometry difference:
\begin{equation}
\label{eq:geo_loss}
\mathcal{L}_g =  \mathbb{E}_{t\sim\mathcal{U}(\{1,\cdots,T\})}\Vert \mathcal{A}_t - \hat{\mathcal{A}} \Vert^2,
\end{equation}
where $\mathcal{L}_g$ is named Angle Summation Loss~(ASL).

\section{Experiments}

\subsection{Experimental Setup}

Our model is evaluated on three standard referring image segmentation datasets: RefCOCO~\cite{yu2016modeling}, RefCOCO+~\cite{yu2016modeling} and RefCOCOg~\cite{mao2016generation}. 
The maximum sentence length $n$ is set 15 for RefCOCO, RefCOCO+, and 20 for RefCOCOg. The images are resized to $640\times 640$. The sampling number of mask vertexes $N$ is set by default to $36$. 
Other data preprocessing operations are generally in line with the previous methods~\cite{zhu2022seqtr}.
As for the total diffusion step $T$, it is set to 1000 during training phrase. During inference phrase, $T$ is set to 4 because DDIM step is adopted for accelerating sampling speed~\cite{song2020denoising}.
Based on previous works \cite{luo2020multi,ding2021vision,zhu2022seqtr}, mask IoU and det accuracy are adopted to evaluate the performance of methods.
AdamW \cite{loshchilov2017decoupled} is adopted as our optimizer, and the learning rate and weight decay are set to 5e-4 and 5e-2.  The learning rate is scaled by a decay factor of $0.1$ at the $60$th step. We train our models for 100 epochs on 4 NVIDIA V100 with a batch size of 64.
All of the quantitative analyses are based on the val split of RefCOCO dataset.

\subsection{Main Results}

\noindent\textbf{Referring Expression Comprehension.} Single-task part of Tab.~{\color{red}{\ref{tab:main_res_det}}} reports the comparison results between our method and previous referring expression comprehension methods.
From Tab.~\ref{tab:main_res_det}, our PVD boosts previous methods by a clear margin.
For example, PVD based on Darknet53~\cite{redmon2018yolov3} respectively surpasses TransVG~\cite{deng2021transvg} and TRAR~\cite{zhu2022seqtr} with +2.74$\sim$+6.13\% and +3.58$\sim$+7.73\% absolute improvement on RefCOCO+.
The results show that our PVD generally achieves SOTA when compared to previous referring expression comprehension methods.
Besides, we construct a stronger network based on Swin Transformer~\cite{liu2021swin} for achieving higher effectiveness, which gets larger improvement than previous methods.

\begin{figure}[t]
\centering
\includegraphics[width=1.0\linewidth]{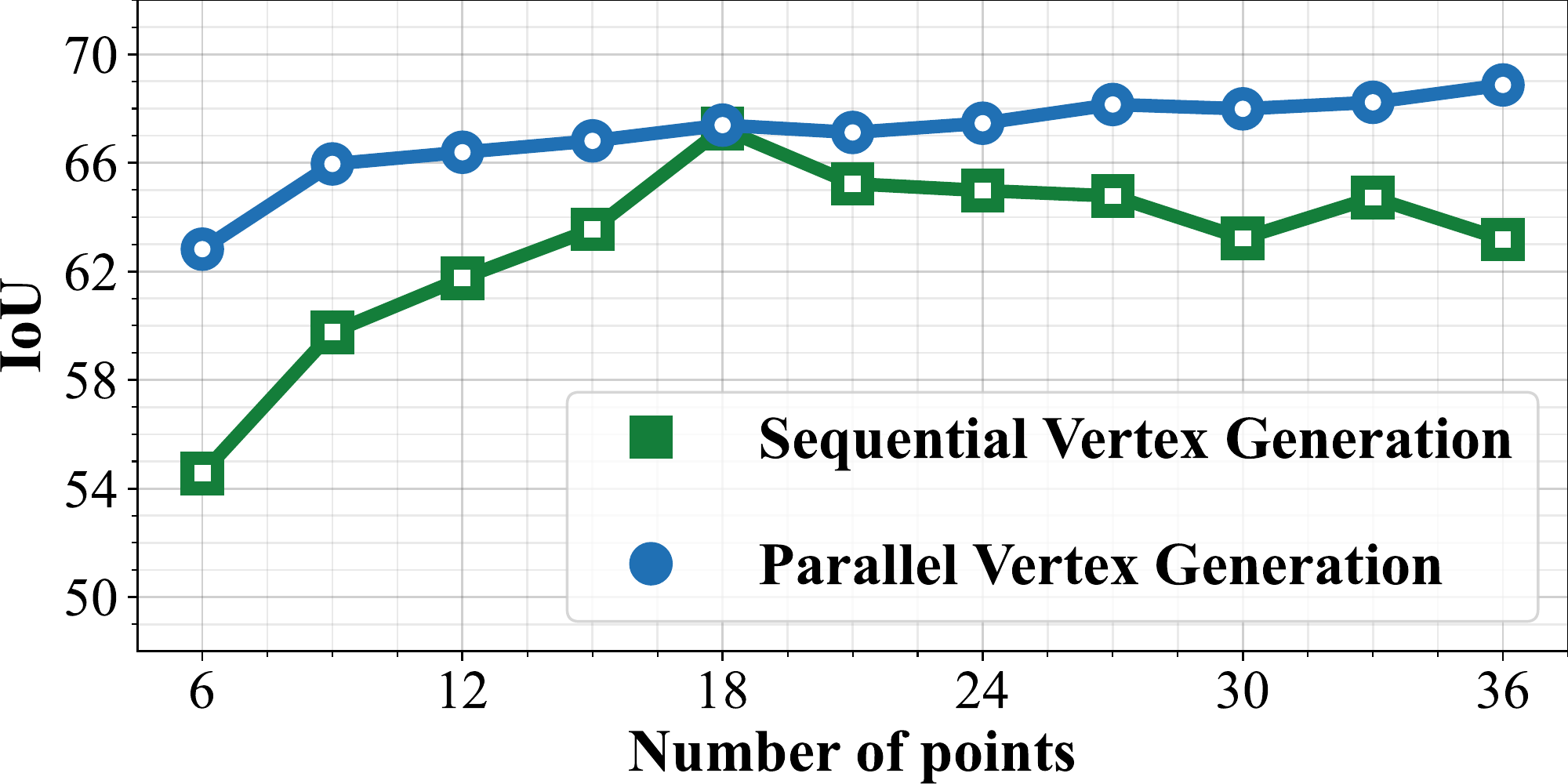} 
% SVG 的不好: 指数时间的序列长度放缩, 误差积累;
\caption{\textbf{Effectiveness Comparison} between sequential vertex generation method~(SeqTR) and parallel vertex generation method~(PVD) with different number of prediction points.
}
\vspace{-5pt}
\label{fig:pt_iou}
\end{figure}
\begin{table}
\centering
\renewcommand{\arraystretch}{1.4}
\setlength{\tabcolsep}{1.5mm}
\footnotesize
\begin{tabular}{x{28}x{28}|y{45}y{45}y{45}}
\noalign{\hrule height 1.5pt}
% \hline
CAM & ASL & Acc (REC) & Acc (RIS) & IoU (RIS) \\
\hline
& & 79.38 & 77.67 & 65.19 \\
\cmark & & 81.24~\increase{1.86} & 79.15~\increase{1.48} & 66.78~\increase{1.59} \\
& \cmark & 81.17~\increase{1.79} & 78.92~\increase{1.25} & 67.31~\increase{2.12} \\
\rowcolor{aliceblue}
\cmark & \cmark & \textbf{82.51~\increase{3.13}} & \textbf{81.03~\increase{3.36}} & \textbf{68.87~\increase{3.68}}\\
\noalign{\hrule height 1.5pt}
\end{tabular}
\vspace{3pt}
\caption{\textbf{Diagnostic Experiments.} ``Acc" of referring expression comprehension~(REC) and referring image segmentation~(RIS) denote the precision@0.5, i.e., the rate of samples with IoU $> 0.5$. ``CAM" denotes the center anchor mechanism proposed in Sec.~\ref{sec:cam} ``ASL" denotes the angle summation loss proposed in Sec.~\ref{sec:asl}.}
\vspace{-5pt}
\label{tab:main_abl}
\end{table}

\noindent\textbf{Referring Image Segmentation.} Single-task part of Tab.~{\color{red}{\ref{tab:main_res_seg}}} reports the comparison results between our method and previous methods.
For comparing to previous SOTA, i.e., vision transformer-based methods~(CoupleAlign, LAVT), we construct a stronger network based on Swin Transformer~\cite{liu2021swin}.
In this case, our PVD can outperform LAVT on all of datasets by +0.8$\sim$+2.09\% and CoupleAlign on most of datasets by +0.12$\sim$1.4\%, which demonstrates our PVD achieves SOTA for referring image segmentation task. 

\noindent\textbf{Unified Visual Grounding.} Multi-task part of Tab.~\ref{tab:main_res_det} and Tab.~\ref{tab:main_res_seg} report the comparison results between our method and previous unified visual grounding methods.
Compared to SOTA method~(SeqTR), our parallel vertex generation paradigm~(PVD w/ Darknet53) outperforms it by +0.74$\sim$+1.46\% for referring expression comprehension and +0.74$\sim$+2.14\% for referring image segmentation, which verifies the superiority of our paradigm.

\subsection{Quantitative Analysis}
\label{subsec:quan_ana}

\noindent\textbf{How does the number of points affect the effectiveness?} 
The main advantage of parallel vertex generation paradigm compared to sequential vertex generation paradigm is easier to scale to high-dimension tasks. To verify this statement, we check the effectiveness of two paradigm with different number of prediction points in Fig.~\ref{fig:pt_iou}. Note that the number of points reflects the dimension of task. The figure provides two justifications: \listnumber{(1)} the performance of sequential vertex generation is bottlenecked at 18 points, which substantiates the \textbf{dimension dilemma} of sequential vertex generation paradigm (the network is perturbed by error accumulation with large number of points and is underfitting to complex object with a small number of points). \listnumber{(2)} the performance of parallel vertex generation stably increases with the number of points, which demonstrates that our paradigm is \textbf{more scalable} to high-dimension tasks.

\noindent\textbf{The efficiency of Parallel Vertex Generation.} Except for effectiveness, efficiency is also an aspect for verifying scalability. To further claim the advantage of parallel paradigm compared to sequential paradigm, we select several point settings~(``9pts", ``18pts", ``27pts", ``36pts") to benchmark the efficiency of two paradigms in Tab.~\ref{tab:pt_iou_speed}. This table shows that the sequential paradigm is heavily impacted by the number of points and needs a large amount of extra computation overheads when scaling from a small number of points to a large number of points. For example, the inference speed becomes 4$\times$ of previous speed when scaling from ``9pts" to ``36pts". However, our paradigm only requires a little extra overhead to scale to high-dimension tasks. Specifically, the inference speed only increases +9ms when scaling from ``9pts" to ``36pts".

\begin{figure}[t]
\centering
\subfigtopskip=0pt
\subfigbottomskip=3.5pt
\subfigcapskip=0pt
\begin{minipage}[b]{0.495\linewidth}
\subfigure[SVG method~(SeqTR w/ 18pts)]{
    \includegraphics[width=0.98\linewidth]{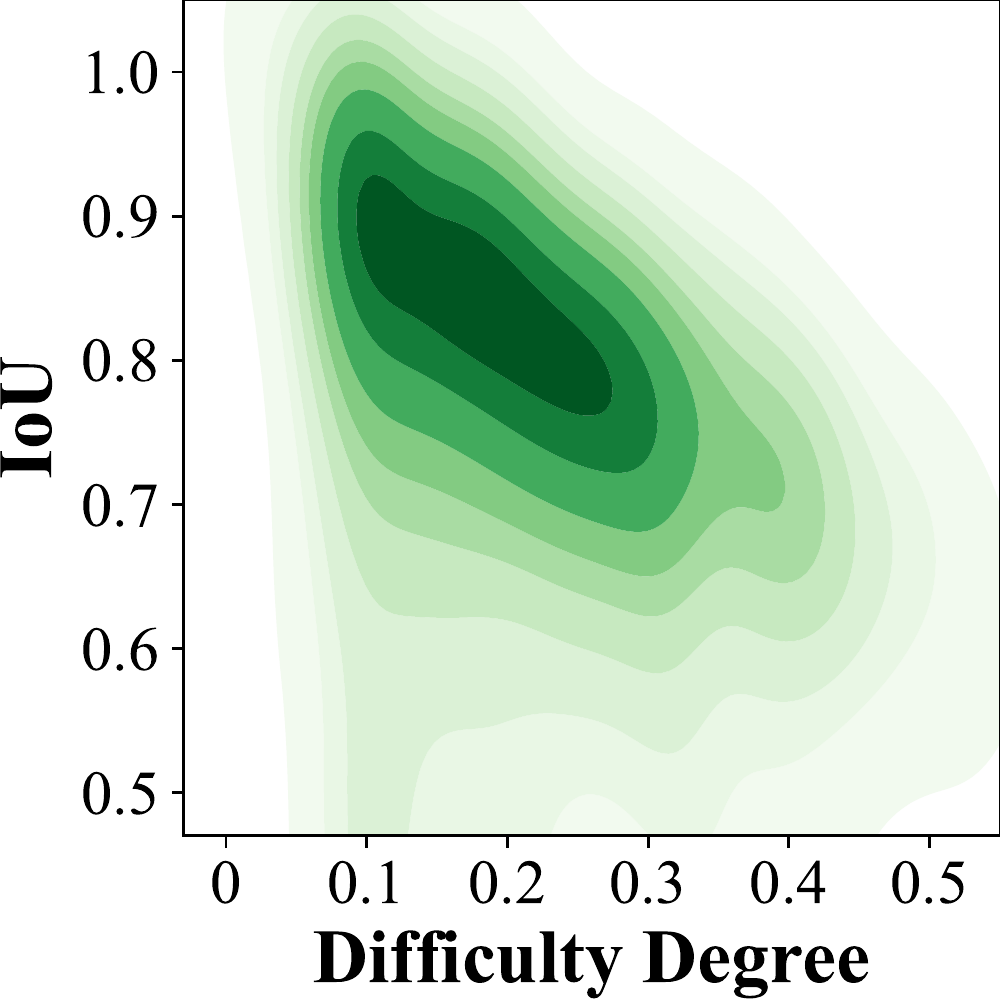}
    \label{fig:diff_iou_dist_svg}
}
\end{minipage}
\hfill
\begin{minipage}[b]{0.495\linewidth}
\subfigure[PVG method~(PVD w/ 36pts)]{
    \includegraphics[width=0.98\linewidth]{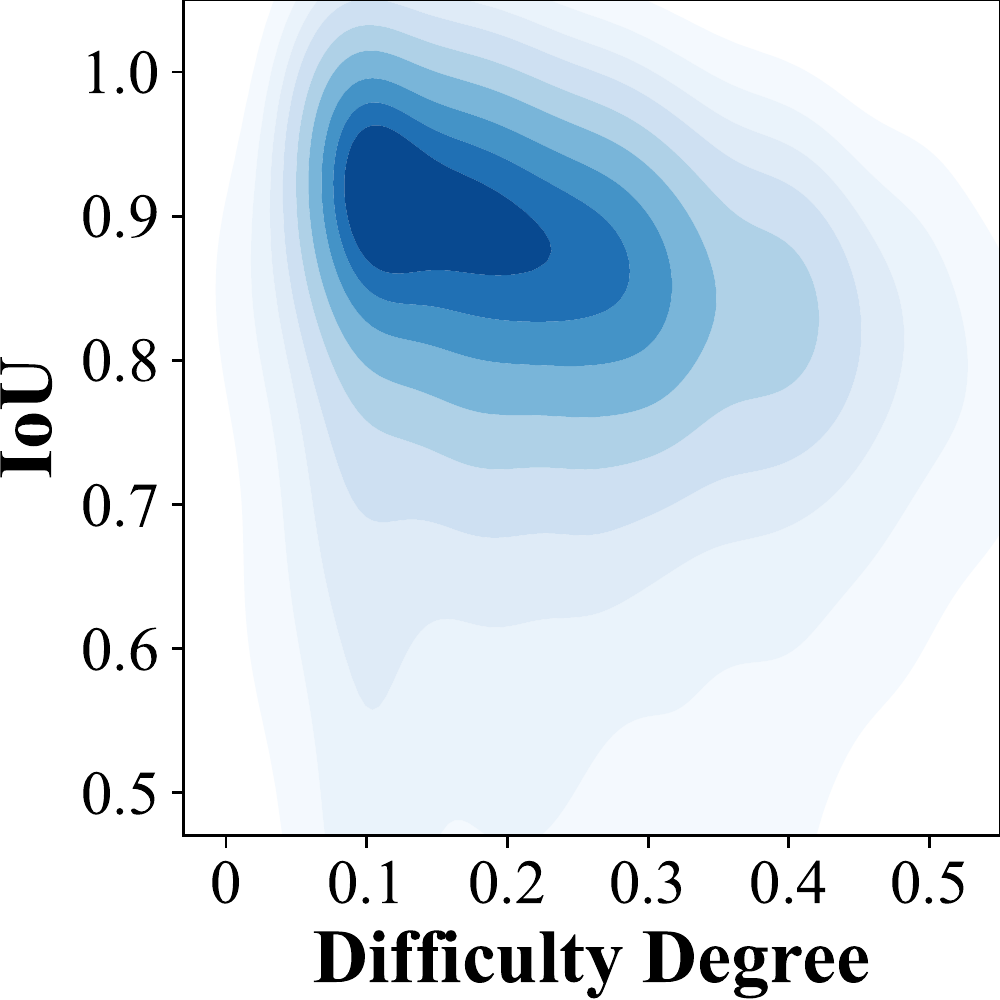}
    \label{fig:diff_iou_dist_pvg}
}
\end{minipage}
\caption{The density map of samples from (a) sequential vertex generation method~(SeqTR) and (b) parallel vertex generation method~(PVD).  Darker area indicates more samples are of the corresponding IoU (\%) value and ``Difficulty Degree". ``Difficulty Degree" denotes the complexity of related object contour.}
\vspace{-5pt}

\end{figure}
\begin{table}
\centering
\renewcommand{\arraystretch}{1.4}
\setlength{\tabcolsep}{1.0mm}
\footnotesize
\begin{tabular}{x{30}|x{44}y{44}|x{44}y{44}}
\noalign{\hrule height 1.5pt}
Number & IoU (SVG) & IoU (PVG) & Speed (SVG) & Speed (PVG) \\
\hline
9pts  & 60.51 & 63.14~\increase{2.63} & 101ms & 66ms~\deincrease{35} \\
18pts & \cellcolor{aliceblue} \textbf{67.26} & 67.39~\increase{0.13} & 192ms & 67ms~\deincrease{125} \\
27pts & 64.78 & 68.15~\increase{3.37} & 285ms & 71ms~\deincrease{214} \\
36pts & 63.18 & \cellcolor{aliceblue} \textbf{68.87~\increase{5.69}} & 394ms & 75ms~\deincrease{319}\\
\noalign{\hrule height 1.5pt}
\end{tabular}
\vspace{3pt}
\caption{\textbf{Efficiency Comparison} between sequential vertex generation~(SeqTR) and parallel vertex generation~(PVD).}
\vspace{-10pt}
\label{tab:pt_iou_speed}
\end{table}

\begin{figure*}[t]
\centering
\includegraphics[width=1.0\textwidth]{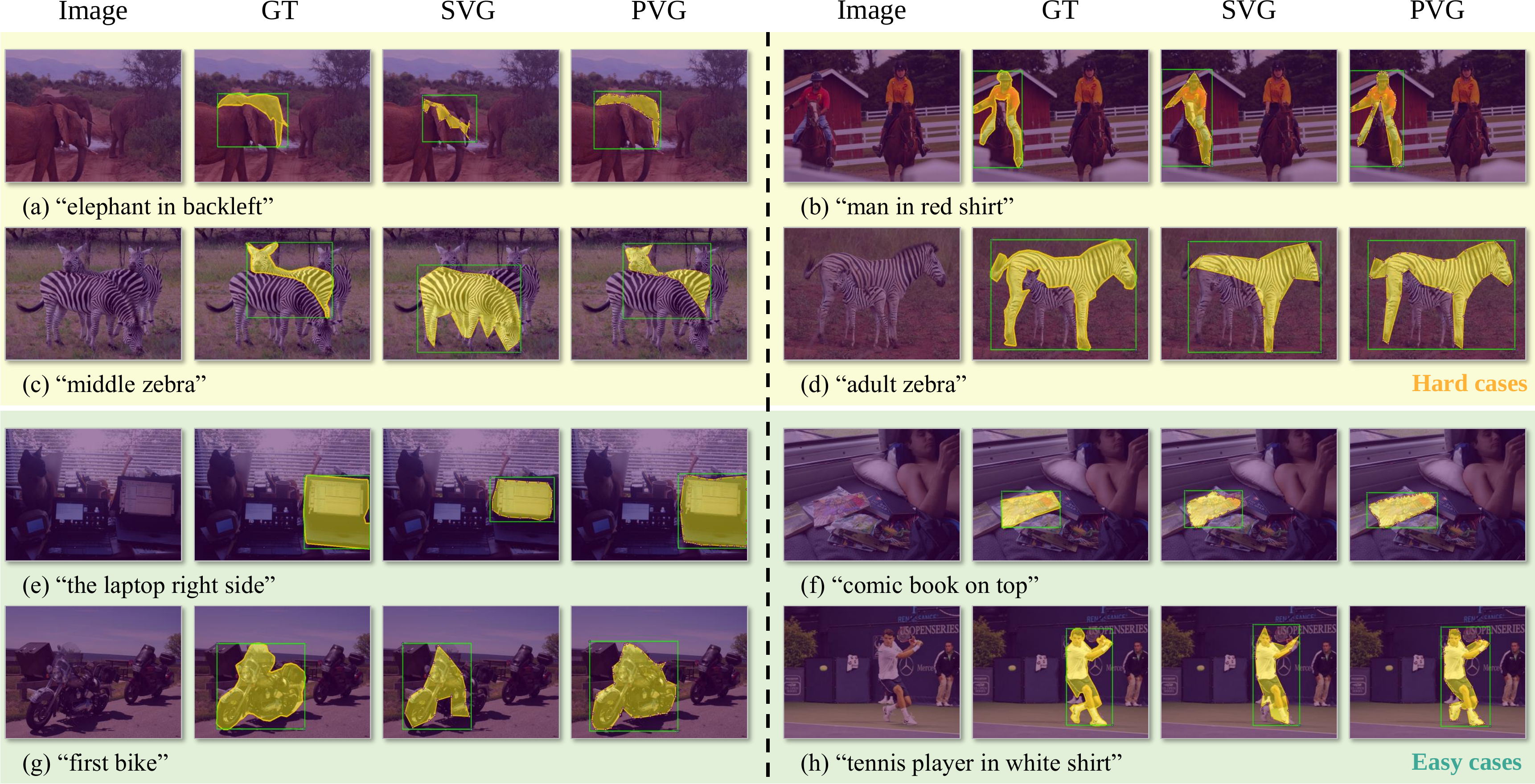}
\caption{\textbf{Qualitative results of different cases.} ``GT" denotes ground truth. ``SVG" denotes SeqTR which follows sequential vertex generation. ``PVG" denotes our proposed PVD which follows parallel vertex generation. We select both hard cases whose referred objects have complex contours and easy cases whose referred objects have monotonous contours for showing the effectiveness of our PVD.}
% , original results of bottom-up branch, original results of top-down branch and the integration results of two branches. There are totally three types of cases selected for showing the effectiveness of our WiCo The first, second and third rows are polar negative cases, inferior positive cases and normal cases.}
\vspace{-7pt}
\label{fig:main_qual}
\end{figure*}

\noindent\textbf{The advantage of scalability.} The scalability ensures that parallel vertex paradigm have favorable effectiveness and efficiency for a large number of points. This advantage makes methods based on our paradigm more robust to samples with complex contours than sequential paradigm. To quantitatively analyze the robustness, we define ``Difficulty Degree" as the complexity metric and count the ``Difficulty-IoU" density map of samples in Fig.~\ref{fig:diff_iou_dist_svg} and Fig.~\ref{fig:diff_iou_dist_pvg}. The calculation details of the metric please refer to appendix. Comparing two figures, we can find that our paradigm still has high IoU for hard samples~(``Difficulty Degree" $> 0.2$) but the IoU of sequential paradigm heavily decreases for hard samples, which justifies that the scalability to high-dimension tasks of our paradigm prompts better robustness to hard samples than seuqnetial paradigm.

\noindent\textbf{The effectiveness of Center Anchor Mechanism.} As mentioned in Sec.~\ref{sec:cam}, CAM is proposed to cope with the high variance of unnormalized prediction vertex coordinates. To verify the effectiveness of CAM, we conduct ablation experiments. Tab.~\ref{tab:main_abl} shows that PVD w/ CAM boosts vanilla PVD by +1.86\% Acc for referring expression comprehension task and +1.59\% IoU for referring image segmentation task, which justifies the effectiveness of CAM.

\noindent\textbf{The effectiveness of Angle Summation Loss.} Since the original training objective of PVD only achieves point-level constraint between prediction and ground truth vertexes, we propose ASL in Sec.~\ref{sec:asl} for geometry constraint. In Tab.~\ref{tab:main_abl}, PVD w/ ASL improves vanilla PVD by +1.79\% Acc for referring expression comprehension task and +2.12\% IoU for referring image segmentation task. Besides, ASL also improves PVD w/ CAM. These results comprehensively verifies the effectiveness of ASL.

\subsection{Qualitative Analysis}

As described in Sec.~\ref{subsec:quan_ana}, our parallel vertex generation paradigm is more effective and efficient than sequential vertex generation paradigm, especially for hard samples. To qualitatively verify this point, we select some easy cases and hard cases to illustrate the grounding difference between two paradigm. Fig.~\ref{fig:main_qual} shows that our parallel vertex generation paradigm generates high-quality vertexes of bounding box and mask contour but the sequential vertex generation paradigm easily hits error objects or generates inferior vertexes, which justifies our paradigm is more capable of grounding referring expression on the image.

\section{Conclusion}

In this paper, we observe that the sequence vertex generation paradigm~(SeqTR) is trapped in a dimension dilemma although it is by far the best paradigm for unified visual grounding. On one hand, limited sequence length causes inferior fitting to objects with complex contours. On the other hand, generating high-dimensional vertex sequences sequentially is error-prone. To tackle this dilemma, we build a parallel vertex generation paradigm to better handle high-dimension settings. Since the diffusion model can scale to high-dimension generation by simply modifying the dimension of noise, it is adopted to instantiate our paradigm as Parallel Vertex Diffusion~(PVD) for pursuing highly scalable unified visual grounding. Subsequently, Center Anchor Mechanism~(CAM) and Angle Summation Loss~(ASL) are designed and introduced into vanilla PVD for improving convergence. Our PVD achieves SOTA on referring expression comprehension and referring image segmentation tasks. Moreover, PVD has a better fitting ability for objects with complex contours and requires fewer computation costs than SeqTR, which justifies our parallel paradigm is more scalable and efficient than sequential paradigm.

{\small
\bibliographystyle{ieee_fullname}
\bibliography{main}
}

\end{document}